\pdfoutput=1

\documentclass[11pt]{article}

\usepackage[preprint]{emnlp/acl}

\usepackage{times}
\usepackage{latexsym}
\usepackage{framed}
\usepackage[T1]{fontenc}

\usepackage[utf8]{inputenc}

\usepackage{microtype}

\usepackage{inconsolata}

\usepackage{graphicx}

\usepackage{microtype}
\usepackage{graphicx}
\usepackage{makecell}
\usepackage{subcaption}
\usepackage{booktabs} 

\usepackage{hyperref}

\usepackage{amsmath}
\usepackage{amssymb}
\usepackage{mathtools}
\usepackage{amsthm}
\usepackage{framed}
\usepackage{listings}

\usepackage{xcolor}
\usepackage{soul}

\usepackage[capitalize,noabbrev]{cleveref}

\usepackage[textsize=tiny]{todonotes}
\usepackage{xhfill}
\usepackage{xspace}

\newboolean{isanonymized}
\setboolean{isanonymized}{false} 

\newcommand{\anonymize}[2]{%
    \ifthenelse{\boolean{isanonymized}}{#1}{#2}%
}

\newcommand{\ppr}{Parametric Proxy Rate\xspace}
\newcommand{\PPR}{PPR\xspace}
\newcommand{\ucr}{Unsupported Correctness Rate\xspace}
\newcommand{\UCR}{UCR\xspace}
\newcommand{\rpa}{Retriever Potential Attainment\xspace}
\newcommand{\RPA}{RPA\xspace}

\title{Quantifying Memorization and Parametric Response Rates in Retrieval-Augmented Vision-Language Models}
\usepackage{authblk}
\author{\textbf{Peter Carragher}\thanks{\textbf{Correspondence:} \href{mailto:petercarragher@cmu.edu}{petercarragher@cmu.edu}}}
\author{\textbf{Abhinand Jha}\thanks{Work done during affiliation with Carnegie Mellon University, now at Google.}}
\author{\textbf{R Raghav}}
\author{\textbf{Kathleen M. Carley}}
\affil{Carnegie Mellon University \\
  Pittsburgh, PA 15217 \\}

\begin{document}
\maketitle

\begin{abstract}

Large Language Models (LLMs) demonstrate remarkable capabilities in question answering (QA), but metrics for assessing their reliance on memorization versus retrieval remain underdeveloped. Moreover, while finetuned models are state-of-the-art on closed-domain tasks, general-purpose models like GPT-4o exhibit strong zero-shot performance. This raises questions about the trade-offs between memorization, generalization, and retrieval. In this work, we analyze the extent to which multimodal retrieval-augmented VLMs memorize training data compared to baseline VLMs. Using the WebQA benchmark, we contrast finetuned models with baseline VLMs on multihop retrieval and question answering, examining the impact of finetuning on data memorization. To quantify memorization in end-to-end retrieval and QA systems, we propose several proxy metrics by investigating instances where QA succeeds despite retrieval failing. In line with existing work, we find that finetuned models rely more heavily on memorization than retrieval-augmented VLMs, and achieve higher accuracy as a result (72\% vs 52\% on WebQA test set). Finally, we present the first empirical comparison of the parametric effect between text and visual modalities. Here, we find that image-based questions have parametric response rates that are consistently 15-25\% higher than for text-based questions in the WebQA dataset. As such, our measures pose a challenge for future work, both to account for differences in model memorization across different modalities and more generally to reconcile memorization and generalization in joint Retrieval-QA tasks. 
\end{abstract}

\section{Introduction} 

The increasing reliance on LLMs for multimodal tasks across far-reaching sectors such as healthcare, finance, and manufacturing underscores the need to assess the accuracy and reliability of the information they generate. Vision-Language Models (VLM) have achieved state-of-the-art (SoTA) performance on Visual Question-Answering (VQA) benchmarks, and these models often utilize Retrieval-Augmented Generation (RAG) to maintain factual accuracy and relevance in a dynamic information environment. However, this has led to uncertainty in the information the LLM bases its answer on in situations where it may choose between parametric memory and retrieved sources. When models rely on memorized information instead of dynamically retrieving information, they may inadvertently propagate outdated or incorrect information, causing serious legal and ethical risks and undermining trust and reliability in AI systems \citep{huang2023survey}.

Despite these concerns, the way that Vision-Language models (VLMs) memorize and retrieve information, particularly in complex multimodal tasks, remains under-explored. Instead, survey studies on parametric knowledge conflicts have found that existing research is focused on reasoning capabilities of unimodal large language models (LLMs) and retrieval augmented generation systems (RAG) \citep{conflicts-main-survey}. Particularly in the context of multimodal retrieval and multihop reasoning, few studies analyze the tradeoff between finetuning for specialized tasks and zero-shot prompting for general-purpose vision-language capabilities. A lack of consensus on how to approach this tradeoff motivates the development of measures to quantify reliance on parametric memory, as well as metrics for quantifying the potential performance impact of extending LLMs with RAG systems.

\begin{figure}
    \centering
    \includegraphics[width=\linewidth]{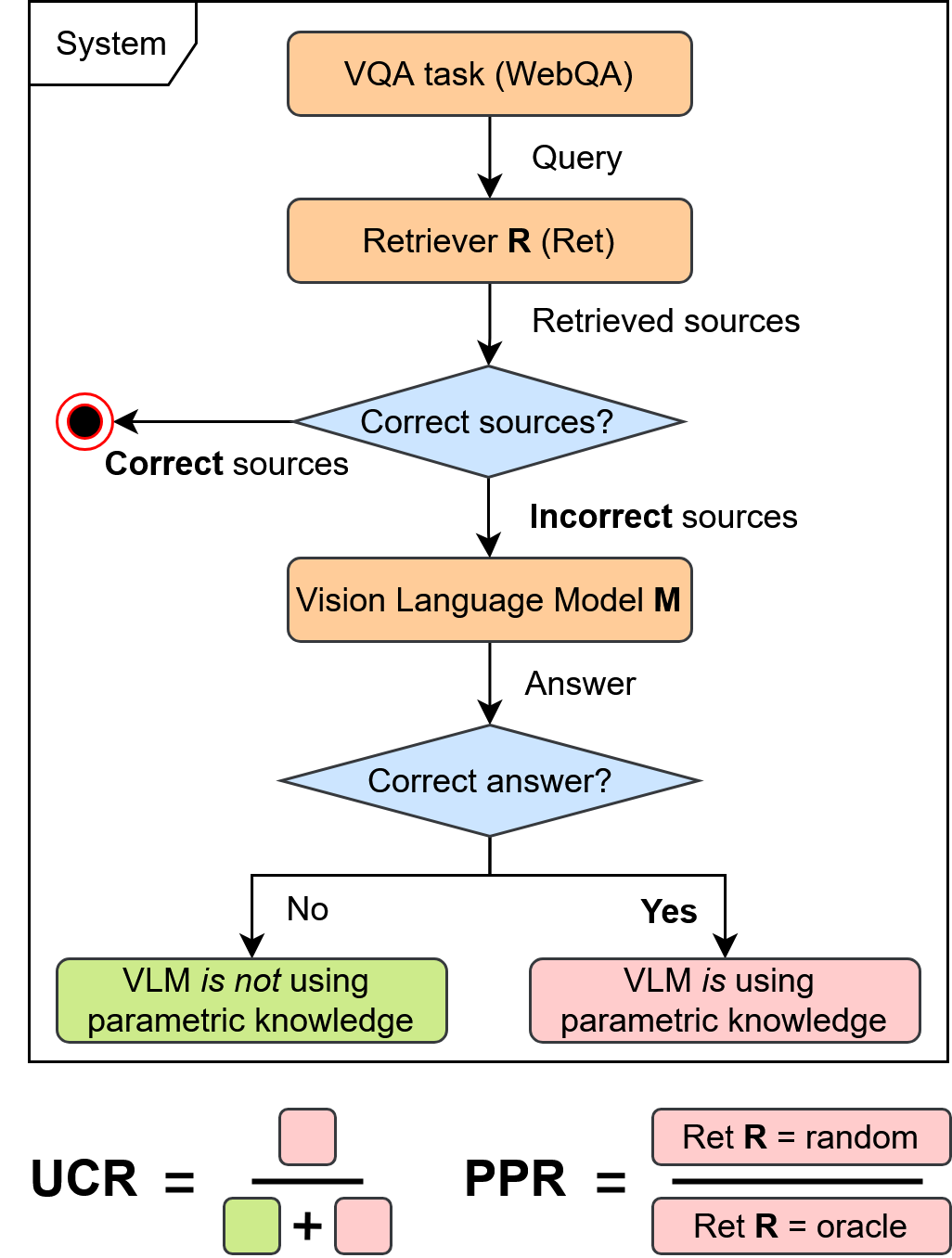}
    \caption{\PPR and \UCR metrics are derived from the interaction between retrieved sources and QA model.}
    \label{fig:overview}
\end{figure}

To address this gap, we investigate how multimodal QA models balance accuracy with memorization on the WebQA benchmark. We compare finetuned multimodal systems against zero-shot VLMs, analyzing how retrieval performance influences QA accuracy. In particular, we focus on cases where retrieval fails, allowing us to measure reliance on parametric memory through two proposed metrics---the \ppr (\PPR) which quantifies how much model accuracy is influenced by retrieval quality, contrasting performance in best-case versus worst-case retrieval scenarios, and the \ucr (\UCR) which measures how often correct QA responses are generated when the retriever fails, providing a proxy for memorization. \autoref{fig:overview} gives an overview of how these measures are derived for a joint retrieval-QA task.

To enable this analysis, we make several methodological contributions. For the finetuned QA models, we investigate Vision-Transformer (ViT) architectures, which allow for multihop reasoning over multiple sources. To investigate the impact of retrieval performance on trained LMs, we propose a variable-input Fusion-in-Decoder (FiD) model \cite{tanaka_slidevqa_2023, nlvr2}, building upon the VoLTA architecture \citep{pramanick_volta_2023}. For the zero-shot case, we build upon previous research on In-Context Retrieval \citep{incontext_rag} by demonstrating that LLMs such as GPT-4o are capable of performing the final ranking step of the retrieval process. In doing so, we find that GPT-4o, a general-purpose LLM, achieves SoTA performance on the WebQA task, outperforming existing finetuned RAG models by a significant margin (7\% higher accuracy). Crucially, our results reveal that while retrieval-augmented models reduce memorization, the training paradigm plays an important role. Finetuned models exhibit higher reliance on parametric memory, whereas zero-shot RAG approaches have lower memorization scores at the cost of accuracy. While retrievers the improve performance zero-shot VLMs to some degree, as is the case for unimodal systems there is yet no silver bullet in the tradeoff between model generalization and specialization.


Finally, we investigate differences in parametric response rates between text-based and image-based questions from the WebQA dataset. We find that models are capable of answering image-based questions based on parametric knowledge 15-25\% more often than they are for text-based sources. In many cases, this means that parametric responses are twice as likely for image-based questions as for text-based ones. Moreover, this result is consistent regardless of the retriever used or whether the QA model was finetuned or not. This finding represents the key empirical contribution from this work---not only is this the first work to measure the parametric effect over image sources, but to the best of our knowledge, it is also the first to present empirical results comparing model memorization tendencies across different modalities. Our findings suggest that the parametric effect may be more pronounced for visual tasks. Future work to validate these findings on additional datasets and problem domains is warranted. We hope that our analysis of model memorization motivates the development of 
transparent and reliable multimodal AI systems, particularly in applications where the sourcing of up-to-date, factual information from multimodal sources is critical.

\section{Related Work} 


\subsection{Multimodal Retrieval Systems} 
A large body of work on multimodal representations exists \citep{liu2022universal,chen_murag_2022,radford2021learning}. 
CLIP enables the embeddings of text and images into aligned representations by supervised training over image-caption datasets \citep{radford2021learning}. More sophisticated local alignment methods between captions and images using Graph Optimal Transport (GoT) have been proposed \citep{chen_graph_2020,maretic_got_2019}. The Universal Vision-Language Dense Retrieval model (UniVL-DR) showed SoTA performance on the WebQA retrieval task \citep{liu2022universal} by using hard-negative sampling for constrastive learning. In this work, we compare UniVL-DR and CLIP embeddings as competing retrieval systems.



\subsection{Multihop Language Models} 
A wealth of research exists on multimodal Vision-Language tasks and multihop language decoders. \cite{tanaka_slidevqa_2023} propose a Fusion-in-Decoder (FiD) architecture for multihop reasoning over images. Utilizing advances in local alignment \citep{chen_graph_2020}, VoLTA model combines graph representations of input questions and source images \citep{pramanick_volta_2023}. For compatibility with retriever modules, we extend VoLTA with support for a variable number of input sequences. 

More recently, the increasing context windows of VLMs enables them to demonstrate multihop reasoning abilities \citep{liu2024visual,abdin2024phi,wang2024qwen2}. Recent work has found that not only are LLMs capable of determining when they should forgo their parametric memory and use a retriever module \citep{labruna2024retrieve}, they are also capable of ``In-context Retrieval" \cite{incontext_rag}. Here, retrieved sources are used for grounded text generation by simply prepending the sources into the input prompt. We expand upon this idea, adapting it to a multimodal setting with VLMs, and report our findings.

\subsection{The Parametric Effect}
There is a wealth of research on reliance on parametric memory for unimodal QA tasks \citep{galway_mitigation_2024,xu_knowledge_2024,longpre_entity-based_2022,neeman_disentqa_2022,hong_why_2024,chen_rich_2022}. Here, the entity replacement framework \citep{longpre_entity-based_2022,neeman_disentqa_2022} is used to invalidate parametric memory by explicitly crafting knowledge conflicts between input sources and parametric memory \citep{xu_knowledge_2024,hong_why_2024,chen_rich_2022}. As such, these studies guarantee that manipulated input sources no longer entail the expected labels, and focus on evaluating LLMs in isolation without using retrieval systems.

In contrast, we do not make the same guarantees, and our proxy measures are premised upon the key assumption that incorrectly retrieved sources \textit{do not entail} the correct answer. Our focus is on developing proxy metrics for the parametric effect that do not require such involved source manipulation processes. Rather, building upon prior work on unimodal LLMs \citep{soudani2024fine}, these metrics compare the performance of finetuned VQA models with RAG systems.

\section{WebQA Dataset} 


The WebQA dataset \cite{chang_webqa_2021} uses a two-step design; retrieval followed by QA. First, given the question Q and all sources $S$, we retrieve the set of relevant sources, $S^\prime$. Using these sources we then generate an answer $A^\prime$. The following is passed to the QA classifier: 
\begin{equation}
    <[CLS], s^\prime_0, [SEP], \dots, s^\prime_n, [SEP], Q, [SEP]>
\end{equation}



We include only those questions that require either one (n = 12,027) or two (n = 9,438) image sources. For a breakdown of question categories and their keywords, see \ref{sec:categories} in the appendix. The remaining questions use only text sources (n = 20,267). Our final analysis of unimodal vs multimodal parametric effects uses this portion of the dataset to evaluate memorization on text sources.

As opposed to WebQA, open-domain VQA tasks such as OK-VQA \citep{marino_ok-vqa_2019} and HotpotQA \citep{yang2018hotpotqa} do not provide candidate sources $S$ and source labels $S^*$, and as a result are incompatible with or measures (see \autoref{sec:measures}). Moreover, while we do evaluate model performance on VQA datasets (NLVR2 \citep{nlvr2} and VQAv2 \citep{goyal2017making}), these tasks lack a retrieval step and so are only useful for QA model selection (see \autoref{sec:vqav2}).



\section{Methodology}

As WebQA is a joint retrieval and QA task, we develop several QA methods and retrieval methods separately. Using the best-performing QA model, we then evaluate end-to-end retrieval and VQA performance and investigate the factors that affect the parametric effect.

\subsection{Question Answering}

\paragraph{Vision-Language Model}
For two-image questions, the WebQA finetuned VLP baseline \citep{zhou_unified_2020} takes as input the concatenation of both sources encodings with the query; 
\begin{equation}
<[CLS], s_1, s_2, [SEP], Q, [SEP]>
\end{equation}

As such, it is an extension of VQA model trained on single-hop VQA-2 \citep{yu_unified_2023}, which takes as input:
\begin{equation}
<[CLS], s, [SEP], Q, [SEP]>
\end{equation}
We adopt this formulation for finetuning the Qwen2 VLM, using Low-Rank Adaptation (LoRA) to reduce trainable parameters \cite{hu2021lora}. We use the same input formulation to evaluate zero-shot performance on GPT-4o. In addition, we evaluate several baseline models from previous works, namely VLP \citep{zhou_unified_2020}, GIT \citep{wang2022git}, GPT-3.5 \citep{brown2020language}, and BLIP-2 \citep{li2022blip}. Details of these models are presented in appendix \autoref{sec:baselines}.


\paragraph{Multihop Formulation}
We hypothesize that multihop tasks, such as WebQA, would benefit from a two-stage reasoning process. The first stage enables multimodal fusion between each input source and the question, and the second stage enables multihop fusion between the embedded multimodal representation of each source, conditioned on the question. Inspired by FiD architectures \citep{yu_kg-fid_2022}, this results in the following input construction:
\vspace{-2mm}
\begin{equation}
\begin{aligned}
    concat(<[CLS], s_1, [SEP], Q>,\\
    <[CLS], s_2, [SEP], Q>)
    \vspace{-1.5mm}
\end{aligned}
\end{equation}

\paragraph{Multihop Classifier}
\label{sec:mh-volta}
We select the VoLTA framework as the skeleton for encoding joint text and image representations \cite{pramanick_volta_2023}. VoLTA uses Swin-Base \cite{liu2021swin} and RoBERTa-Base \cite{liu2019roberta} as respective visual and textual encoders and we adopt the same encoder choices. 
We jointly encode each image source returned by the retriever with the query and concatenate the resulting embeddings together before sending them to the MLP classifier to predict the keyword answer label. To handle variable input sequences during classification, we pad single image sources with blank images so that all inputs sent into the classifiers have two images. We call this model MultiHop-VoLTA (MH-VoLTA).

We finetune the models using the AdamW \cite{loshchilov2019decoupled} optimizer with a learning rate of $1e^{-4}$ and a batch size of 32 samples. We use LoRA to reduce trainable parameters \cite{hu2021lora}, and set $r=8$ and $\alpha=32$ for the text encoder and $r=16$ and $\alpha=16$ for the image encoder, updating only the attention weights. MH-VoLTA is trained until convergence (~80 epochs, see \autoref{fig:loss_convergence} in the appendix). 


\subsection{Retrieval Methods}
\label{sec:retrieval_models}
\paragraph{Dense Retrievers} We adopt the pretrained UniVL-DR retriever for source retrieval in our finetuned experiments \citep{liu_universal_2023} and compare it with baseline CLIP \citep{radford2021learning} and WebQA finetuned CLIP (CLIP-DPR, \cite{liu_universal_2023}) embeddings. Specifically, we embed all text sources, image sources, and queries using UniVL-DR. For each query, we compute cosine similarity between the query and each of the sources, and use the top two ranked image sources and their captions as input to the QA model.

\paragraph{GPT-4o Ranking} 
We utilize GPT-4o to select sources from the set of distractor sources present in dataset using the prompt in the appendix \autoref{frame:labeling_prompt}. This is motivated by previous work in In-Context Retrieval Augmented Language Modeling (In-Context RALM) which demonstrated that LLMs are capable of reasoning over sources without finetuning \citep{incontext_rag}.

\paragraph{Upper and Lower Bounds}
In addition, to investigate the impact of the parametric effect on joint retrieval and QA performance, we also compare performance with a best and worst case retriever. The best case is the oracle retriever, using gold sources provided in the validation set, and the worst case is a random naive retriever, which returns random distractor sources (and so is always incorrect).

\section{Evaluation Metrics} 
We propose measures for evaluating the degree of memorization in QA models (\ppr) and in end-to-end retrieval-QA systems (\ucr), as well as a metric for retriever-QA model compatibility (\rpa).

\label{sec:measures}
\subsection{\ucr}
We propose \UCR, a metric to measure the parametric effect in the combined retrieval and QA model. It is formulated as a composition of QA accuracy and retrieval recall. Intuitively, it is the fraction of true positive predictions from the QA model for which there is no retrieval support (i.e. the retrieved sources were incorrect).

\paragraph{Retrieval Recall and QA Accuracy}
The first stage of the joint task is retrieval, where the recall for retriever R is defined as the fraction of retrieved sources (positives) that are correct (true positives) with respect to task labels;
\begin{equation}
    \text{Recall}_R = \frac{\text{True Positives}}{\text{True Positives} + \text{False Positives}}
\end{equation}

Accuracy is the primary correctness metric for question answering in the WebQA task. Accuracy of model M is determined by comparing a restricted bag of words (bow) vector between the expected (E) and generated (G) answers;
\begin{equation}
    \label{eq:ACC}
    \text{Acc}_M = \frac{1}{n}\Sigma [\frac{|\text{bow}_{E} \cap \text{bow}_{G}|}{|\text{bow}_{E}|} == 1]
\end{equation}

The vocabulary of the vectors is restricted to a specific domain based on the question type; questions are labeled based on these domains which can be yes/no, color, shape, or number. Each category has a pre-defined vocabulary list, given in the \hyperref[sec:categories]{appendix}.





\paragraph{\UCR}
Using QA accuracy and retrieval recall, we construct \UCR, a metric for measuring the parametric effect in a combined retrieval-QA model, which calculates the likelihood $P(Q_1|R_0)$ that the QA model M returns a correct answer ($Q_1$) given that the retrieval model R failed to return the correct sources ($R_0$):
\begin{equation}
\label{eq:UCR}
\begin{aligned}
\text{UCR(R,M)} &= \frac{\text{Acc}_M == 1 \cap   \text{Recall}_R == 0}{\text{Recall}_R == 0}\\
&= P(Q_1|R_0) 
\end{aligned}
\end{equation}

\subsection{Oracle-Normalized Retrieval Scores}
Using min-max scaling, we define two additional metrics to evaluate joint retrieval QA systems, by normalizing using the oracle retriever (upper bound) and random retriever (lower bound):
\begin{equation}
\hat{X} = \frac{X - X_{min}}{X_{max} - X_{min}}
\end{equation}

\paragraph{\rpa}
\RPA quantifies the potential that a retriever has realized when used in a given end-to-end retrieval QA system. The upper bound (1) is given by same QA system's accuracy with oracle sources ($\text{Acc}_M(\text{oracle})$). The lower bound (0) is given by the random negative source retriever ($\text{Acc}_M(\text{random})$), which always retrieves incorrect sources. Note that $\text{Acc}_M(\text{random})$ is similar to the Free Success Rate metric \citep{lin2022retrieval}, which represents the QA model's accuracy with no retrieved sources.
We apply random-oracle scaling, where $Acc_M(R)$ denotes the accuracy of QA model M, given sources from retriever R: 
\begin{equation}
\label{eq:RPA}
\text{\RPA}(\text{R,M}) = \frac{\text{Acc}_M(R) - \text{Acc}_M(\text{random})}{\text{Acc}_M(\text{oracle}) - \text{Acc}_M(\text{random})}
\end{equation}

\paragraph{\ppr}
We postulate that the rate at which a model's performance increases when used in conjunction with increasingly accurate retrievers implies that it is using the retrieved sources effectively, instead of relying on parametric memory. To that end, we present \PPR, an additional max scaling measure based off just the upper (oracle) and lower bound (randomized \textit{negatives}) retrievers, which is simply their performance ratio with respect to QA model M:
\begin{equation}
\label{eq:PPR}
\text{\PPR}(\text{M}) = \frac{\text{Acc}_M(\text{random})}{\text{Acc}_M(\text{oracle})}
\end{equation}


\section{Results} 
        

        
        

First, we experiment with multiple QA models to determine which generalized LLMs and specialized, finetuned models should be selected for the joint retrieval and QA task. Then, we analyze how performance on the retrieval task impacts QA accuracy and experiment with different combinations of retrieval systems and chosen QA models for the joint task. Finally, we investigate the effect of finetuning on model memorization and compare parametric response rates over WebQA image-based and text-based questions.


\subsection{Model Selection}
 We find that the \hyperref[sec:mh-volta]{MH-VoLTA} model outperforms all baseline and zero-shot models on the WebQA validation set image questions, including \hyperref[sec:blip]{BLIP-2}, \hyperref[sec:GIT]{GIT}, \hyperref[sec:VLP]{VLP}, GPT-4o, and \hyperref[sec:gpt3.5]{GPT-3.5}. We also find that MH-VoLTA performance is comparable to VoLTA on the (fixed input) VQA and NLVR2 tasks (section \ref{sec:model_selection} in the appendix). For a breakdown of model performance by question category on the WebQA dataset, see section \ref{sec:category_perf} in the appendix. 

\begin{figure}[!t]
    \centering
    \begin{subfigure}{\linewidth}
        \includegraphics[width=\linewidth]{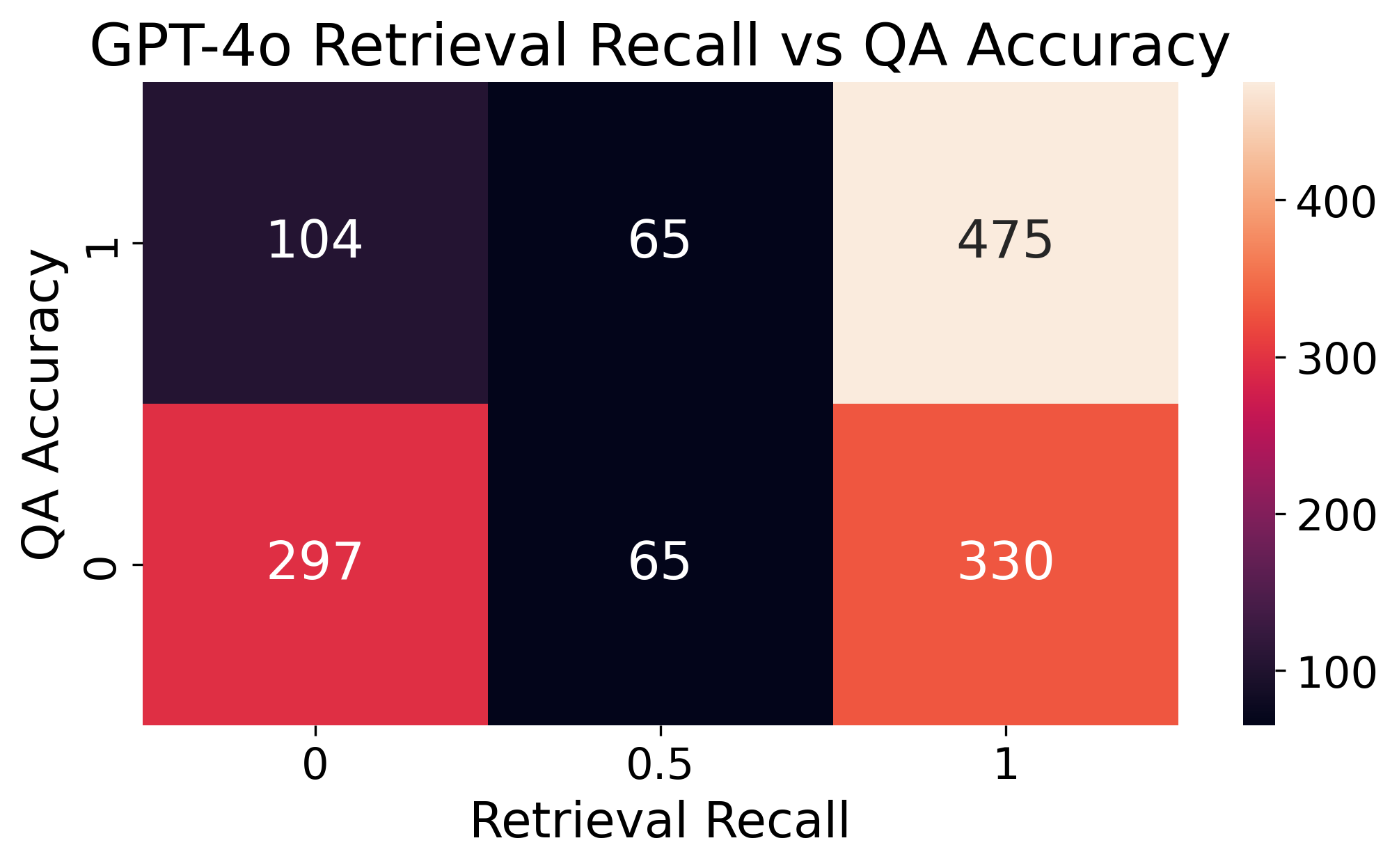}
        \caption{Most questions answered correctly by GPT-4o have correctly retrieved sources, as given by low UCR scores: 
        \(\ensuremath{\UCR(\text{GPT},\text{GPT}) = \frac{104}{104+297} = 0.26}\)}
        \label{fig:retrieval_vs_qa_heatmap}
    \end{subfigure}
    
    \begin{subfigure}{\linewidth}
        \includegraphics[width=\linewidth]{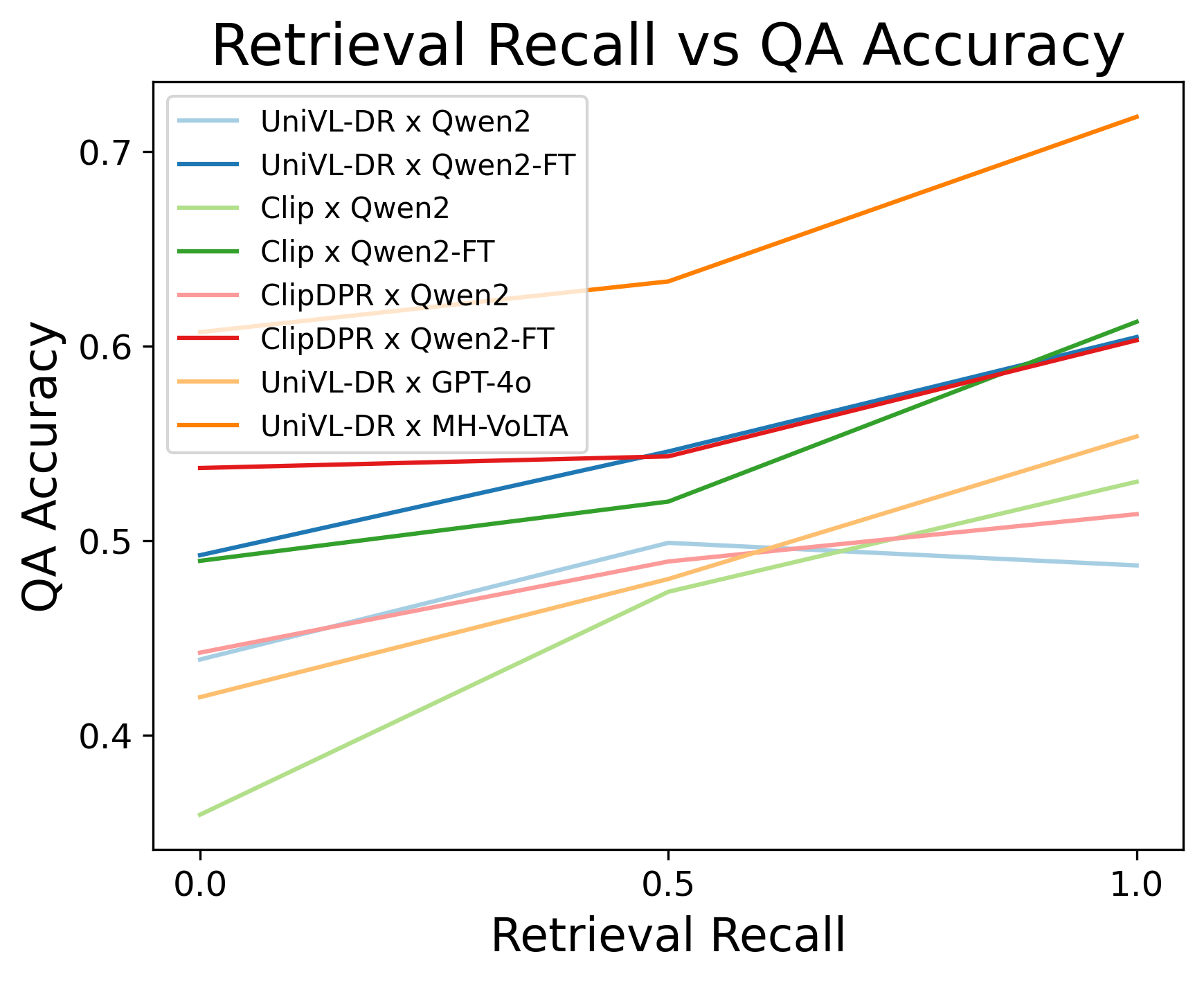}
        \caption{Retrieving distractor sources decreases QA accuracy.}
        \label{fig:retrieval_vs_qa_lines}
    \end{subfigure}
    \caption{Across experiments, recall impacts QA.}
    \label{fig:retrieval_vs_qa}
\end{figure}

\subsection{Impact of Retrieval on QA}
To understand how retrieval and QA systems interact, we investigate the reliance of the QA task on retrieval correctness (\autoref{fig:retrieval_vs_qa}). We find that when GPT-4o correctly retrieves the relevant sources through In-Context Retrieval, it has a 59\% QA accuracy rate. If GPT-4o In-Context Retrieval fails to retrieve the correct sources, the accuracy rate is reduced to 26\% (\autoref{fig:retrieval_vs_qa_heatmap}). We also find that QA performance drops as the number of retrieved distractors increases and retrieval recall falls, showing that poor retrieval performance adversely affects QA (\autoref{fig:retrieval_vs_qa_lines}). This is to say that the QA task is heavily dependent on retrieval performance. However, there do exist correctly answered questions for which incorrect sources are retrieved, and these samples form the basis for the \UCR measure of the parametric effect (\autoref{fig:retrieval_vs_qa_heatmap}).

In addition, we contribute error analysis in the appendix that show the differences between In-Context Retrieval using GPT-4o and dense retrieval methods such as UniVL-DR (\autoref{appendix:complexity_analysis_metrics}). In particular, we find that systems that rely on "in-context" retrieval using GPT-4o are limited by query complexity, but approaches that utilize dense retrievers are not. As query complexity increases, In-Context Retrieval degrades QA accuracy (\autoref{fig:complexity_vs_qa}), but dense retrievers do not (\autoref{fig:complexity_vs_qa_ft}).


    

\subsection{End-to-End Retrieval and QA}
\begin{figure}
    \centering
    \includegraphics[width=\linewidth]{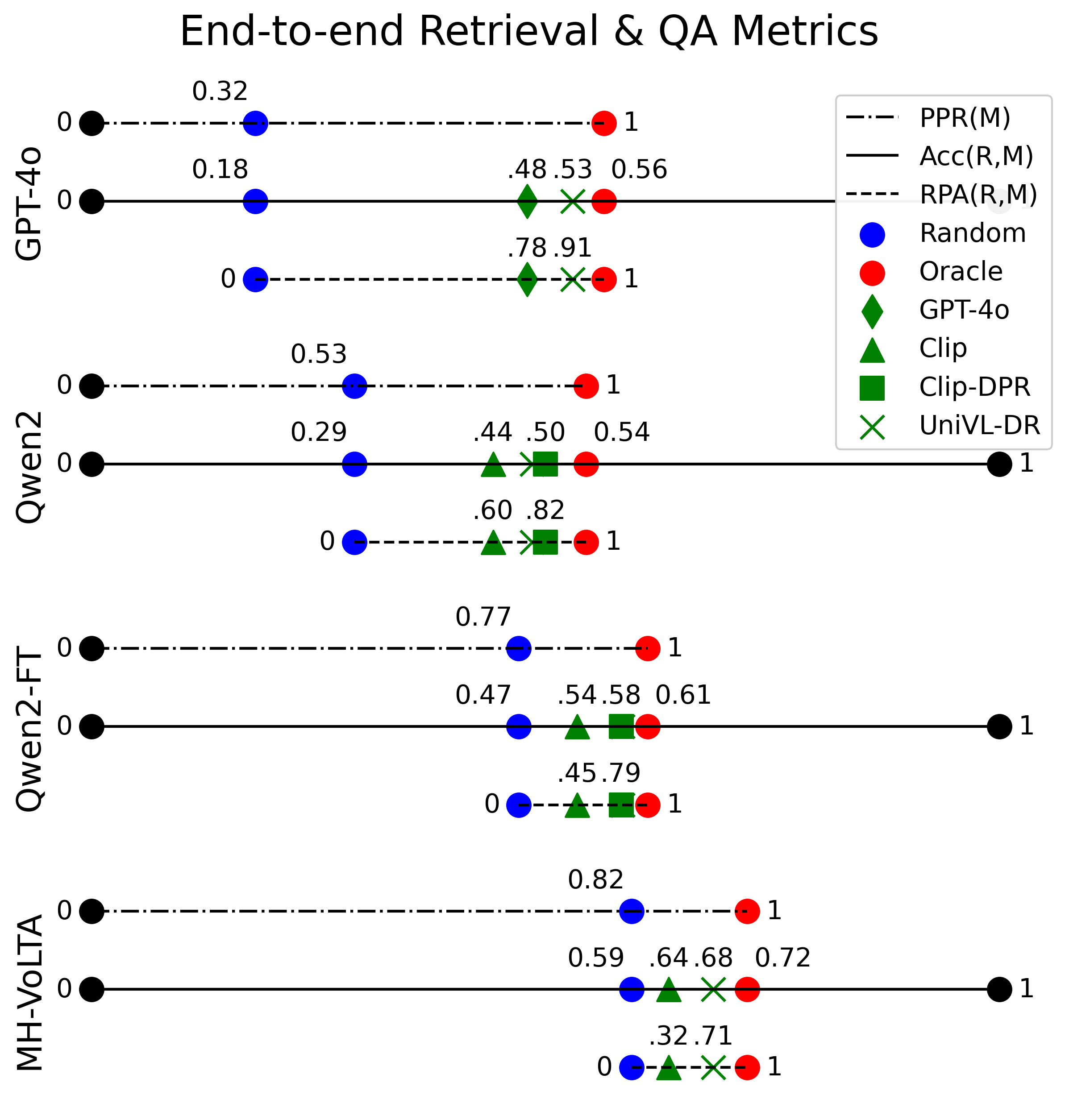}
    \caption{Evaluation metrics on end-to-end retrieval \& QA systems: Accuracy (Acc, \autoref{eq:ACC}), \ppr (\PPR, \autoref{eq:PPR}), and Retriever Attainment (\RPA, \autoref{eq:RPA}) (denoted by the three lines) for each pairing of retriever R and QA model M. Note that the lines denoting \PPR and \RPA are rescaled to represent the denominators in the respective equations. }
    \label{fig:eval_lineplots}
\end{figure}

The \PPR and \RPA measures enable a quick comparison of joint retrieval and QA systems, where \autoref{fig:eval_lineplots} reveals some interesting trends. We find that of all QA models tested, GPT-4o benefits the most from the use of retrievers---\rpa (\RPA) scores are highest for GPT-4o---while finetuned QA models such as Qwen2-FT and MH-VoLTA receive a lower performance increase as the coupled retrieval system is improved.

GPT-4o also has the best \PPR score. That is, GPT-4o has the biggest gap in performance when comparing the worst case (random negative) and best case (source oracle) retrievers, with a \PPR of 0.32. In comparison, Qwen2 has higher performance under the random retriever, and as such displays a greater reliance on parametric memory. 

There is also a clear trend between finetuning, QA accuracy, and \ppr (\PPR). While finetuned Qwen2 (Qwen2-FT) has improved accuracy vs Qwen2, it's performance on the worst case retriever is surprisingly high (\PPR=0.77). This is even more extreme for MH-VoLTA, which obtains both the highest QA accuracy (0.72) and the highest \PPR (0.82). The same trend is apparent when evaluating on the WebQA test set, where finetuning Qwen2 improves accuracy (\autoref{tab:test_scores}). Note, \PPR cannot be measured on the test set, as labels have not been made public.

\begin{table}
    \centering
    \begin{tabular}{ccr}
    \toprule
Retriever R & Model M & $\text{Acc}^{\text{test}}_M(R)$ \\
\midrule
UniVL-DR & Qwen2 & 0.52 \\
UniVL-DR & Qwen2-FT & 0.70 \\
Uni-VLDR & GPT-4o & 0.73 \\
GPT-4o & GPT-4o & \textbf{0.77} \\ 
\bottomrule
    \end{tabular}
    \caption{QA Accuracy on WebQA test set.}
    \label{tab:test_scores}
\end{table}


\subsection{Finetuning and \UCR}
We find that, while the finetuning process improves accuracy (and in part because of this fact), finetuning exacerbates the parametric effect. Qwen2-FT has a higher \ppr than the baseline Qwen2 model (\PPR, \autoref{fig:eval_lineplots}), and it has a higher \ucr (\UCR) than Qwen2 across all retrieval methods tested (\autoref{tab:ucr_scores}). What's more, the act of finetuning Qwen2 has an outsized effect on \UCR when compared with the effect that changing the retriever has. MH-VoLTA represents the extreme case; for each retriever R, $\text{\UCR}(R,\text{MH-VoLTA})>0.5$, implying that MH-VoLTA is correctly answering the majority of questions for which the retrieval system fails to identify the correct sources. 

However, the effect of retrieval on \UCR is not negligible, and we find that for a given QA model, \UCR increases as retrieval recall increases; i.e. for each model M $\text{\UCR}(\text{Rand}, M)<\text{\UCR}(\text{Clip}, M)<\text{\UCR}(\text{ClipDPR}, M)$ (\autoref{tab:ucr_scores}). This implies that as the retriever improves, the QA model is more successful on samples that retrieval fails on. This paradox is explained by inaccuracies in the source labels---distractor sources often provide enough context for the QA model to answer correctly. Rather than exposing memorization, this reveals an underlying issue with the source labels in the WebQA dataset, and as such, these measures can be adapted to evaluate the correctness of the joint retrieval-QA benchmarks.

\begin{table}[!t]
\centering
\begin{tabular}{clllll}
\toprule
Retri- & recall & \multicolumn{4}{c}{\ucr} \\
\cline{3-6}
ver R & & Qwen & Q-FT & MHV & GPT4 \\ 

\midrule
Rand & 0.00 & 0.260 & 0.449 & 0.595 & 0.174 \\
Clip & 0.46 & 0.328 & 0.467 & 0.617 & -- \\
CDPR & 0.77 & 0.420 & 0.517 & 0.643 & -- \\
UniVL & 0.80 & 0.438 & 0.521 & 0.616 & 0.420 \\
GPT4 & 0.65 & -- & -- & -- & 0.259 \\
\bottomrule
\end{tabular}
\caption{\UCR ($P(Q_1|R_0)$) for each retriever R and QA model M, alongside retrieval recall. CDPR denotes Clip-DPR, Q-FT is Qwen2-FT.}
\label{tab:ucr_scores}
\end{table}

\subsection{Multimodal vs Unimodal Memorization}
\begin{figure*}[!h]
    \centering
    \begin{subfigure}{0.3\textwidth}
      \centering
      \includegraphics[height=5.7cm]{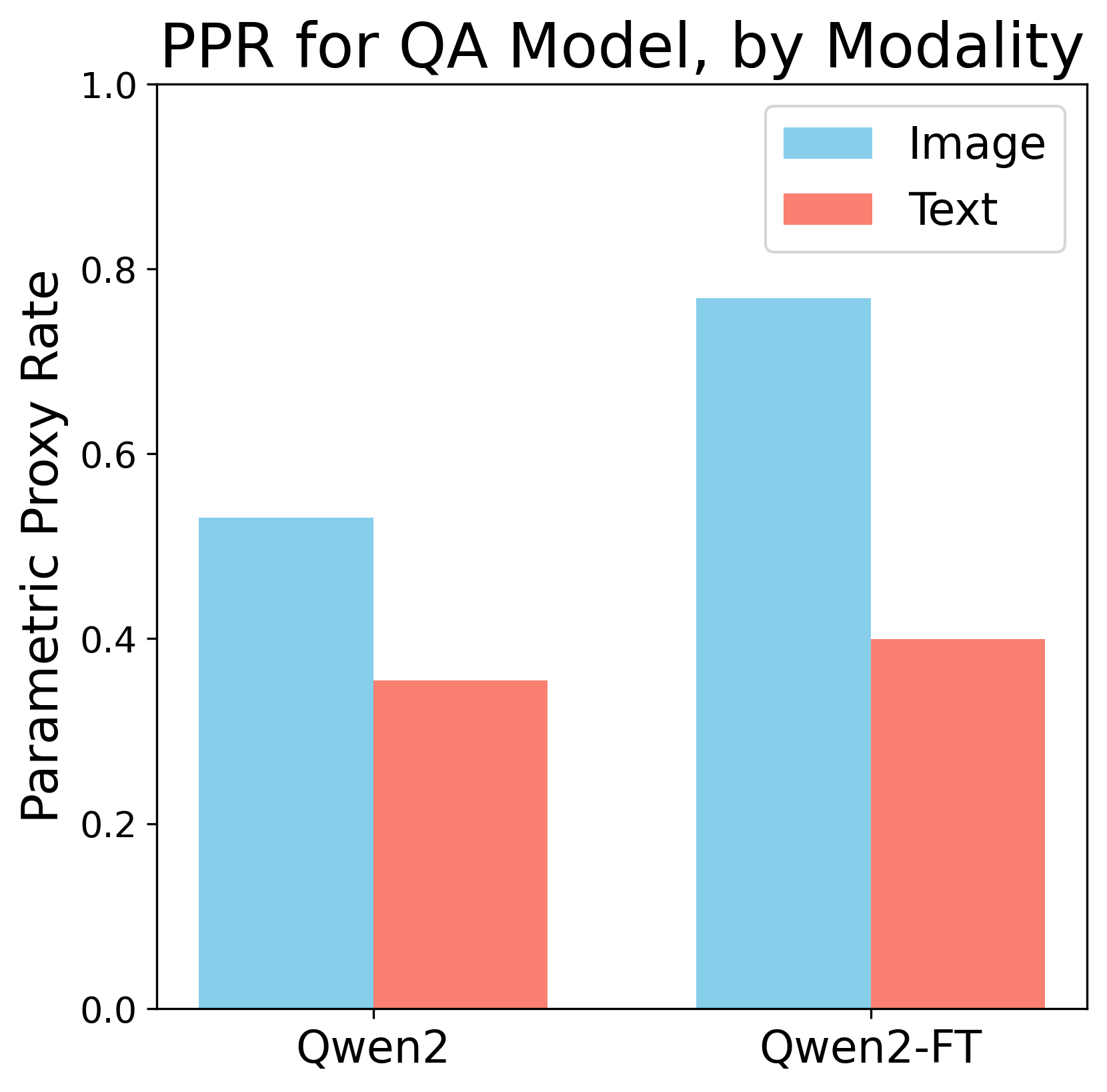}
      \caption{$\text{\PPR}(\text{image}) > \text{\PPR}(\text{text})$}
      \label{fig:ppr_modality}
    \end{subfigure}%
    \hfill
    \begin{subfigure}{0.63\textwidth}
      \centering
      \includegraphics[height=5.7cm]{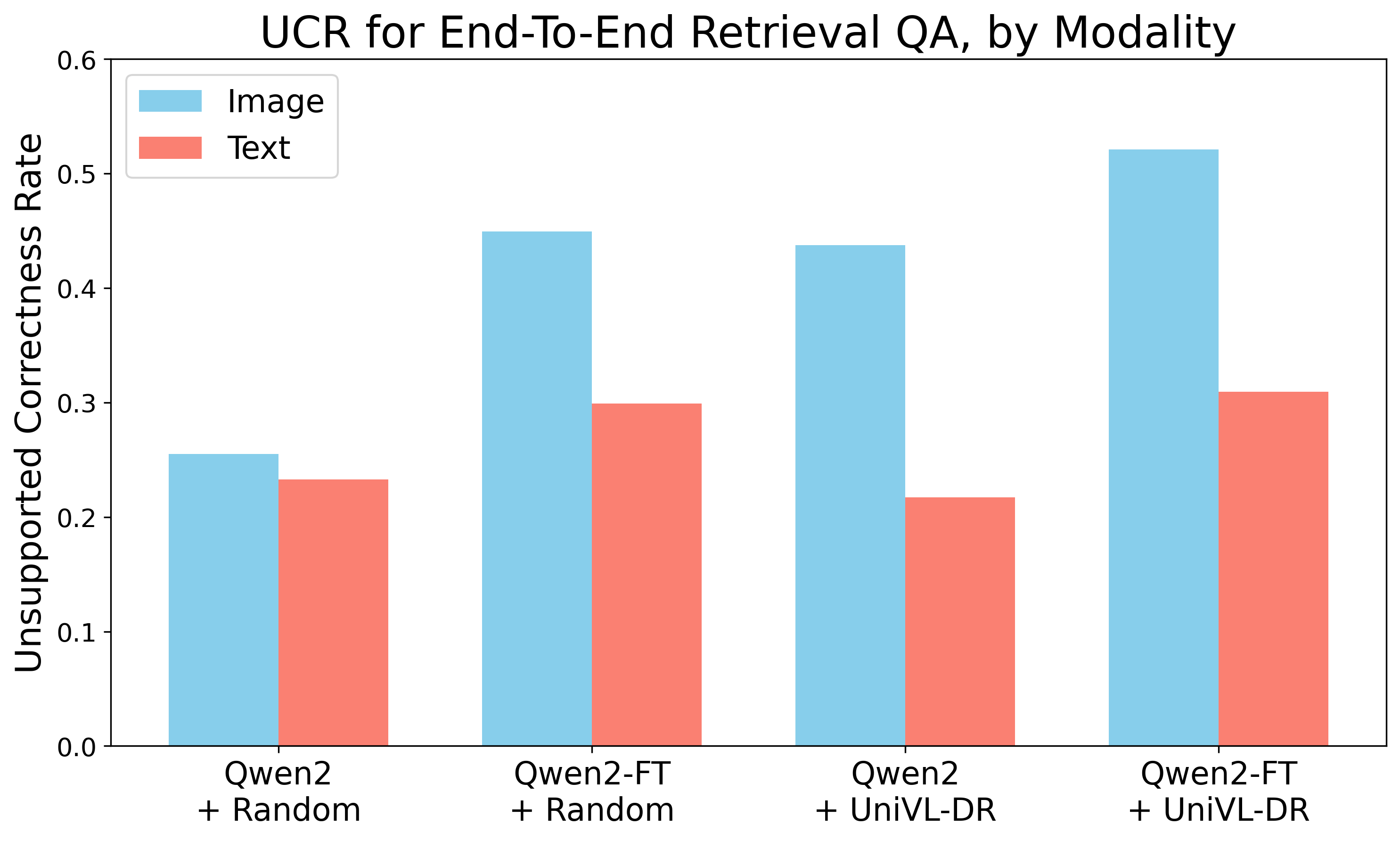}
      \caption{$\text{\UCR}(\text{image}) > \text{\UCR}(\text{text})$}
      \label{fig:ucr_modality}
    \end{subfigure}
\caption{Higher implies greater levels of memorization. Across all metrics, QA models, and retrievers tested, parametric responses are more prevalent for webqa image questions than for webqa text questions.}
\label{fig:modality}
\end{figure*}

Finally, using the developed metrics and the finetuned and non-finetuned VLM models, we investigate the differences between textual and visual parametric knowledge. \autoref{fig:modality} reveals that parametric responses are more prevalent for webqa image questions than for webqa text questions. While finetuning results in a minor increase of parametric response rates for the text modality, these rates increase dramatically for the visual modality after finetuning. For example, while UCR on text-based questions increases by $\sim10\%$ with finetuning, it increases by $\sim20\%$ for image-based questions. In addition, parametric responses are as much as twice as likely when models are presented with image sources, and these results are consistent across the metrics and models used. For example, for Qwen2-FT, $\text{\PPR}_{img} = 0.77$ while $\text{\PPR}_{txt} = 0.4$, and when Qwen is paired with the UniVL-DR retriever, the end-to-end system has $\text{\UCR}_{img} = 0.44$ and $\text{\UCR}_{txt} = 0.22$. These results highlight a modality-based disparity in how memorization manifests in QA models.

\section{Discussion}
While QA performance is generally predicated upon retrieval success (\autoref{fig:retrieval_vs_qa_lines}), there are many cases where retrieval fails and QA succeeds (\autoref{fig:retrieval_vs_qa_heatmap}). These cases form the basis of our quantitative metrics, the \ucr (\UCR; see \autoref{tab:ucr_scores}) and \ppr (\PPR; see \autoref{fig:eval_lineplots}), and with these measures we show that external retrievers significantly reduce the reliance of VLMs on parametric memory. This not only preserves model flexibility but also mitigates the over-specialization common in finetuned systems. However, despite GPT-4o obtaining state-of-the-art performance on the WebQA benchmark using this approach (\autoref{tab:test_scores})\footnote{``Anon\_Feb25" @ \href{https://eval.ai/web/challenges/challenge-page/1255/leaderboard/3168}{WebQA leaderboard}}, for less powerful VLMs such as Qwen2 the decrease in \ppr associated with not finetuning the model (\PPR: 0.77 -> 0.53) comes at the cost of model accuracy (QA accuracy: 70\% -> 52\%).




In interpreting these results, it is important to note that \UCR and \PPR are proxy measures based upon the key assumption that incorrectly retrieved sources \textit{should} result in incorrect answers from the VQA model. While we can guarantee that distractor sources from the random retriever provide no useful information to the model (i.e. they are randomly sampled negatives), fully addressing this assumption requires modifying the image sources to invalidate the original answer or label. Along these lines, a recent line of research uses constrained image generation to create knowledge conflicts between image sources and parametric memory \citep{carragher2025segsub}. This builds on research into provoking parametric responses from unimodal LLMs, where the entity replacement methods \citep{longpre_entity-based_2022,neeman_disentqa_2022} create knowledge conflicts between text sources and parametric memory \citep{xu_knowledge_2024,hong_why_2024,chen_rich_2022}. Entity replacement frameworks for VLMs \citep{carragher2025segsub} can make use of object detection \citep{ravi_sam_2024} and visual attention models \citep{selvaraju_grad-cam_2020} to allow parametric analysis to move beyond the incorrectness assumption.

Despite this assumption, our measures reveal an interesting interplay between retrieval and parametric responses. By providing insights into end-to-end retrieval and QA systems, \UCR can highlight when models are over-reliant on parametric memory. High Retriever Attainment scores (\RPA) across Qwen2 and GPT-4o experiments demonstrate that general-purpose VLMs can utilize finetuned retrievers (\autoref{fig:eval_lineplots}), drawing the need for domain-specific finetuning into question. This work points towards In-Context Retrieval as a particularly promising direction for future research in multimodal systems, if the limitation regarding question complexity can be addressed (\ref{appendix:complexity_analysis_metrics}). 

Crucially, while image manipulation methods are not subject to the incorrectness assumption that the proxy metrics proposed here are, they are not applicable to text sources. Our proxy measures allow for the comparison of how the parametric effect manifests across different modalities in multimodal language models. Herein, we find that parametric responses may be more prominent over image sources as opposed to text sources (\autoref{fig:modality}). Given that the parametric effect is, as of yet, understudied in multimodal setting \citep{conflicts-main-survey}, to the best of the authors knowledge this is the first comparison of parametric effects between modalities. Our finding motivates incorporating the wealth of parametric research on unimodal models into the multimodal domain \citep{carragher2025segsub}.

Overall, our methodology provides a framework for measuring and mitigating memorization in retrieval-augmented systems, offering new ways to evaluate the quality of retrieval-QA datasets. Future work can apply this framework to improve retrieval-aware finetuning strategies, where LLMs learn when to prioritize retrieved content rather than rely on parametric knowledge \citep{labruna2024retrieve}. Extending our parametric analysis to open-domain multimodal tasks would provide new insights into retrieval dynamics in unrestricted, real-world settings. Finally, by quantifying reliance on parametric memory, researchers can better assess the trade-offs within finetuning and retrieval per modality, thereby guiding the development of multimodal models that balance generalization with task-specific performance. As retrieval-augmented VLMs continue to scale, our findings highlight the need for multimodal evaluation of parametric responses to ensure safe, effective, and adaptable AI systems.

\subsection{Limitations}
The \UCR analysis presented in \autoref{fig:ucr_modality} is subject to limitations based on the design of the metric. Specifically, if certain question categories have higher retrieval recall, the number of incorrectly retrieved samples for that category will be low. A study of \UCR across question categories, including suitable confidence intervals, is warranted.

Our analysis also reveals a paradoxical relationship between retriever recall and \UCR that highlights a potential annotation issue within the dataset, prompting a reevaluation of how retrieval-QA benchmarks are constructed (\autoref{tab:ucr_scores}). While this means that \UCR can be used in the evaluation of retrieval tasks to highlight potential false negative annotation issues, as in WebQA (\autoref{tab:ucr_scores}), annotation inconsistencies in turn affect the reliability of \UCR as a sole indicator of retrieval quality. As such, our findings should be validated on additional VQA datasets. To facilitate this, widely used VQA datasets could be augmented with retrieval tasks, such that joint retrieval-QA systems may be evaluated on them.

Finally, given the rapid pace of research in the multimodal space, the WebQA dataset may have already been incorporated into the training data of the VLMs investigated here. While WebQA is not specifically listed among the Qwen2-VL training materials \citep{wang2024qwen2}, many similar datasets are. For models that have been pretrained on the same joint retrieval-QA dataset used to operationalize the measures proposed here, the measures may be more indicative of verbatim memorization, where model output exactly matches the dataset labels. In contrast, our analysis is targeted at the parametric effect, which is a preference for models to reason over parametric knowledge instead of input sources. Disentangling verbatim memorization from the parametric effect is a good avenue for future research, as it represents a more dramatic failure case of model generalization.

\section{Conclusion} 
We demonstrate that retrieval-augmented VLMs have improved performance over general-purpose VLMs, with comparable parametric response rates. However, there is still a substantial performance gap between finetuned and baseline QA models. By introducing UCR and PPR, we provide concrete measures of how incrementally improving retrieval systems mitigates parametric responses. This analysis outlines the interplay between parametric knowledge and external retrieval, highlighting well-known tradeoff between memorization and generalization in the multimodal setting. Finally, we demonstrate that current VLMs have higher parameteric response rates when reasoning over image sources rather than text sources. Our work provides a foundation for future research aimed at refining retrieval mechanisms and ensuring that external sources effectively complement the parametric knowledge of VLMs.

\newpage
\appendix

\section{Appendix}


\subsection{Model Selection Results}
\label{sec:model_selection}
We explore baseline methods for the QA task on the WebQA validation set. \autoref{tab:baselines} gives results for the baseline models. The MH-VoLTA model outperforms all baseline and zero-shot models on the validation set image questions. However, the extension of the VoLTA model for variable input multi-hop tasks risks a regression in performance on traditional VQA tasks which have fixed-input where the number of input images is constant. To determine MH-VoLTA generalizes from fixed to variable input tasks, we compare performance between two variants of the original VoLTA model, finetuned on one and two image subsets of WebQA, with MH-VoLTA. We find that MH-VoLTA is capable of reasoning over both one and two-image image questions, and it's performance is on-par with VoLTA variants trained on one and two image sources separately\autoref{tab:baselines}. See \autoref{sec:one_vs_two_image_volta} for more details on the one and two image VoLTA variants, as well as a breakdown of model performance by question category (\autoref{fig:multihop_volta_res}). See \autoref{sec:baselines} for a description of the baseline models used.



\paragraph{VQAv2 and NLVR2}
\label{sec:vqav2}

As our memorization metrics require that the task be designed in a two-part retrieval and VQA process, this leaves WebQA as the only valid VQA task to evaluate performance on. Here, independently of any external retrieval system, we validate MH-VoLTA performance on two fixed-input VQA datasets (see \autoref{tab:baselines}). As the other baseline models have been validated in prior works, we do not measure their performance on VQAv2 or NLVR2 again.

We evaluate models VQAv2 \citep{goyal2017making}, a multi-class, single-image VQA dataset, and \citep{nlvr2}, a binary classification, two-image VQA dataset. These datasets are well-suited to VoLTA classifier architecture. In particular, question categories in VQAv2, along with the associated answer-domains, match well with WebQA, with a substantial portion of both datasets focusing on color, shape, number, and yes/no questions.

\subsection{Multihop VoLTA on one vs two image sources}
\label{sec:one_vs_two_image_volta}
\begin{figure*}
    \centering
    \begin{subfigure}[b]{0.3\textwidth}
        \includegraphics[width=\linewidth, trim={0 0 0 3cm},clip]{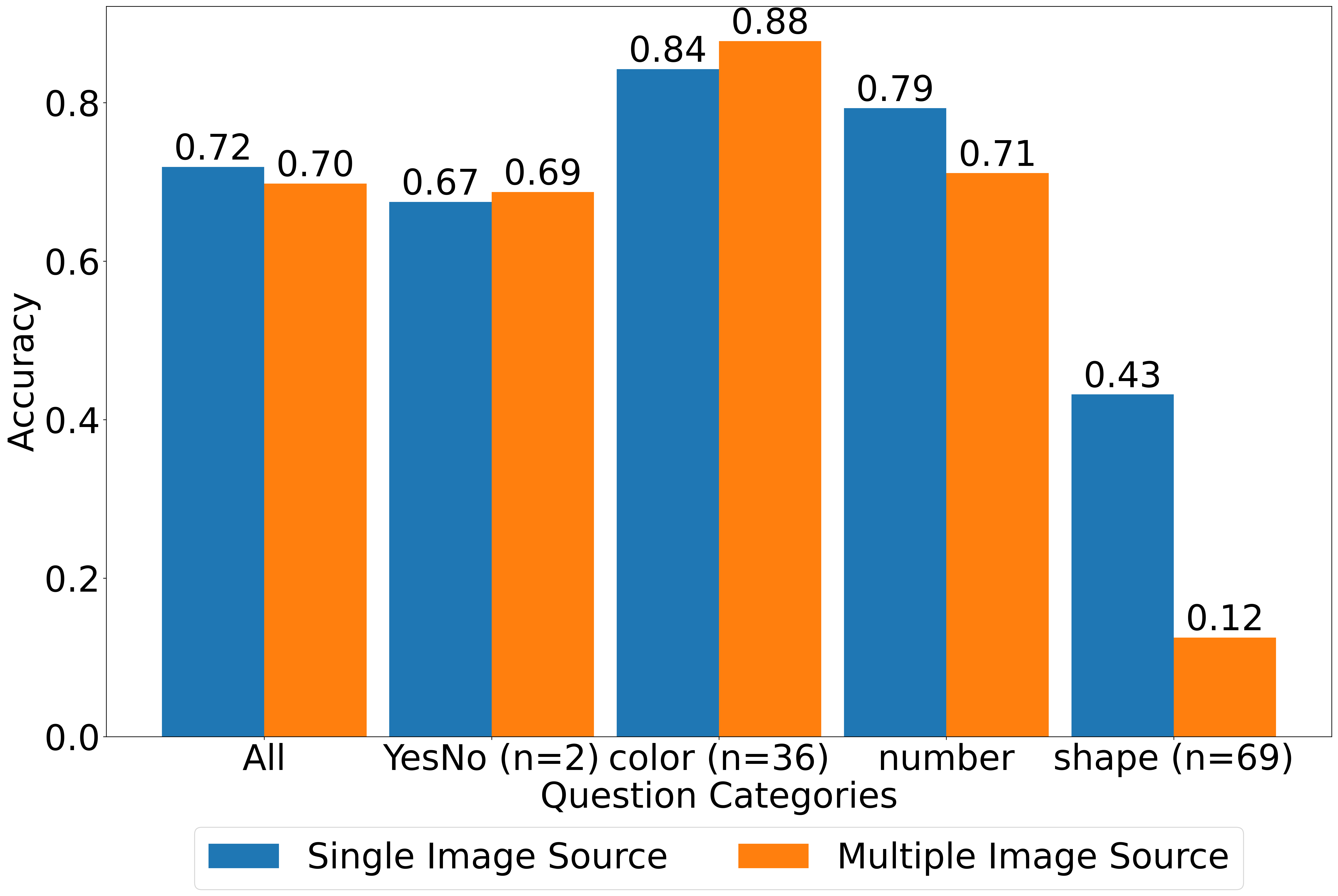}
        \caption{Performance of the MH-VoLTA classifier by question category and image count.}
        \label{fig:multihop_volta_res}
    \end{subfigure}\hfill
    \begin{subfigure}[b]{0.3\textwidth}
        \includegraphics[width=\linewidth]{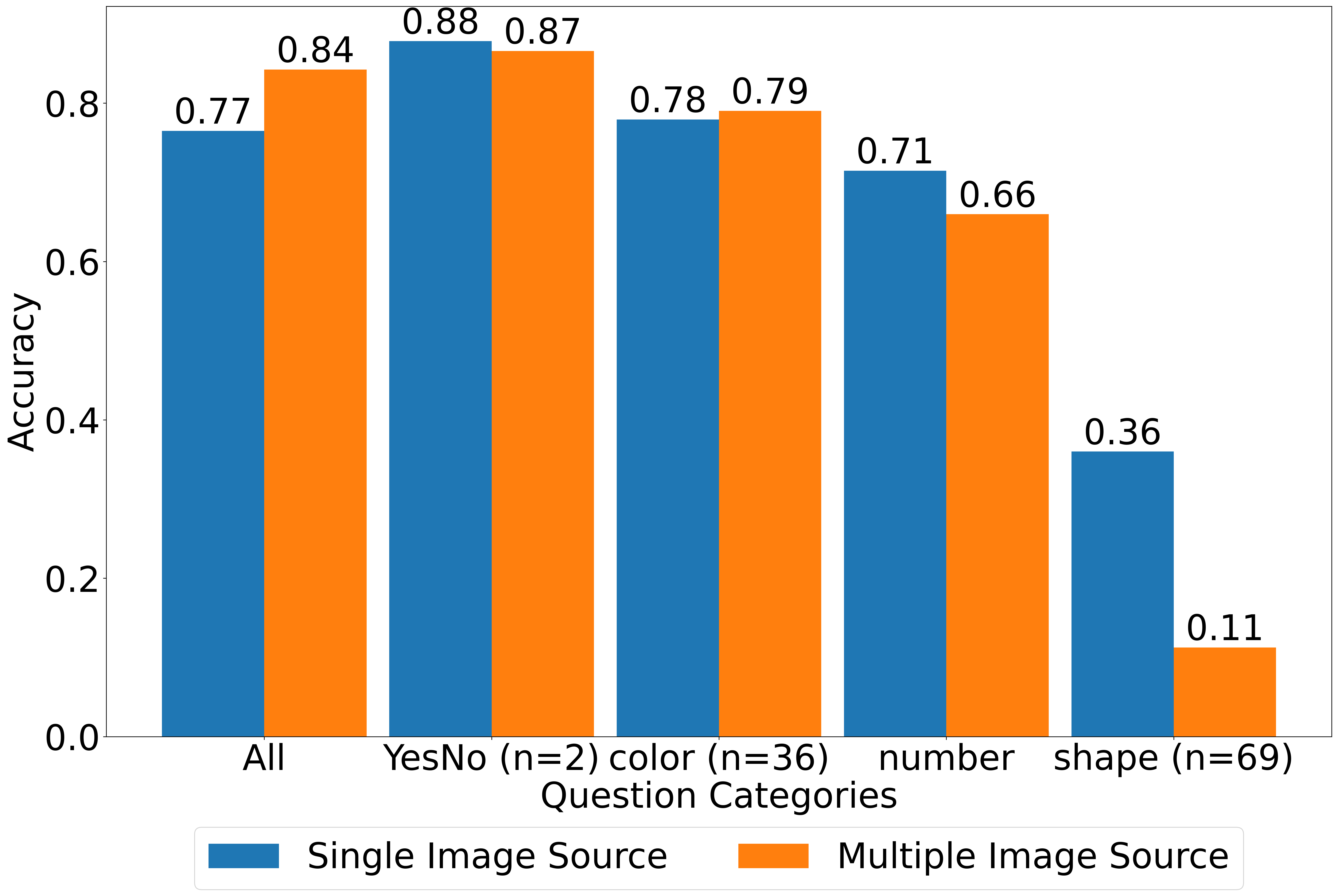}
        \caption{Performance of fixed input single and two-image VoLTA classifiers.}
        \label{fig:model_perf_qcate}
    \end{subfigure}\hfill
    \begin{subfigure}[b]{0.33\textwidth}
        \includegraphics[width=\linewidth]{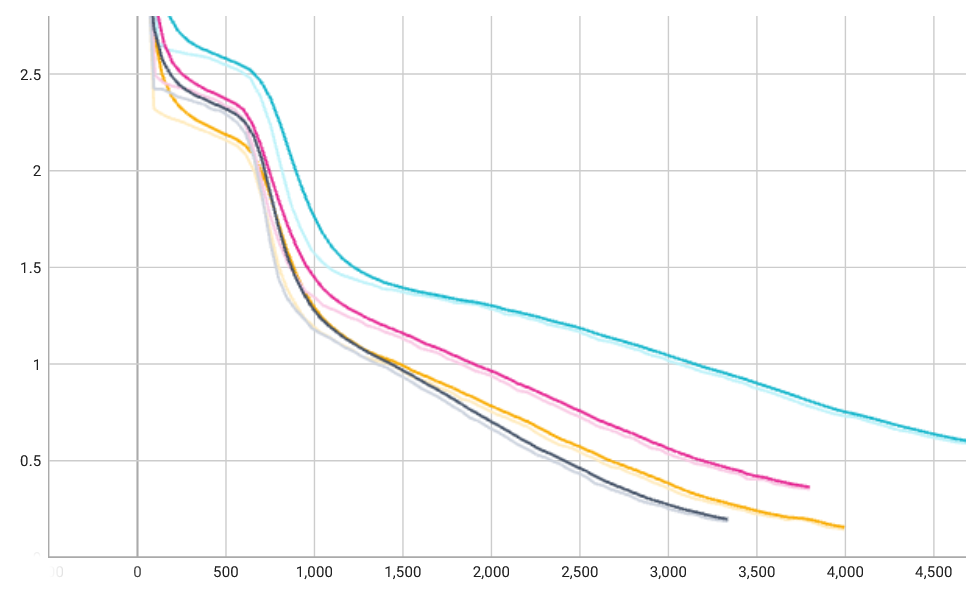}
        \caption{Convergence of the VoLTA loss function on the WebQA dev set across several MH-VoLTA training runs.}
        \label{fig:loss_convergence}
    \end{subfigure}
    \caption{Comparison of the variable MH-VoLTA model (left) vs fixed input VoLTA models (center) across different question categories, ordered by the number of image sources per question. Models converge after ~80 epochs (right).}
\end{figure*}

The results for finetuning VoLTA and MH-VoLTA on the WebQA dataset experiments are provided in \autoref{tab:multihop_volta_res}. We explored the application of Multihop-VoLTA in addressing queries based on single images, questions involving two images, and a combination of both single and two-image queries (referred to as multiple images, \autoref{fig:model_perf_qcate}). 

We find that the variable Multihop-VoLTA model (\autoref{fig:multihop_volta_res}) is en-par with the fixed-input one and two-image VoLTA model variants (\autoref{fig:model_perf_qcate}). This underscores the stability of our finetuning approach for MH-VoLTA across both training paradigms. The MH-VoLTA models have on the order of 100M parameters, of which 10M are trainable after applying LoRA. All models are trained for 80 epochs on a Nvidia A6000.



\begin{table}
    \centering
    \caption{Model selection results on WebQA validation set (further broken into 1 and 2 image input categories), and the VQAv2 and NLVR2 (NLV) test sets. MH-V denotes MH-VoLTA. See \autoref{sec:baselines} for model descriptions.}
    \label{tab:baselines}
    \begin{tabular}{clllll}
    \toprule
     & \multicolumn{3}{c}{\makecell{WebQA Acc}} & VQA & NLV \\
        \midrule
         Model & All & 1 img & 2 img & Acc & Acc \\
         \midrule
         \hyperref[sec:mh-volta]{MH-VoL} & \textbf{0.71} & 0.72 & 0.70 & 73.9 & 76.5 \\
         VoLTA\textsubscript{1}  & -- & \textbf{0.77} & -- & \textbf{74.6} & -- \\
         VoLTA\textsubscript{2}  & -- & -- & \textbf{0.84} & -- & \textbf{76.7} \\
         \midrule
         GPT-4o & 0.56 & \textbf{0.58} & 0.69 &  -- & --\\ 
         Qwen2 & 0.54 & -- & -- &  -- & --\\ 
         \hyperref[sec:gpt3.5]{GPT-3.5} & 0.53 &  0.41 & 0.45 & -- & --\\ 
         \hyperref[sec:VLP]{VLP} & 0.50 & 0.40 & 0.42 & -- & -- \\ 
         \hyperref[sec:GIT]{GIT}  & 0.42 & 0.43 & 0.35 & -- & --\\ 
         \hyperref[sec:blip]{BLIP-2} & 0.40 &  0.37 & 0.44 & -- & --\\ 
         \bottomrule
    \end{tabular}
\end{table}

\begin{table}
    \centering
    \caption{MH-VoLTA results and dataset breakdown}
    \label{tab:multihop_volta_res}
    \begin{tabular}{ccc}
    \toprule
         & No. of Samples&Accuracy\\
         \midrule
         Single Image& 760&0.764\\
         Two Images& 576&0.851\\
         Multiple Images& 1336&0.799\\
         \bottomrule
    \end{tabular}
\end{table}

\subsection{Performance by Question Category}
\label{sec:category_perf}
We report the mean accuracy per question category for Multihop-VoLTA in \autoref{fig:multihop_volta_res} using source retrieval oracles. We find that performance is dependent upon the level of training data available, with the shape category having the least number of samples in the dataset. Question counts per category are as follows; Yes/No (n = 7,320), color (n = 1,830), number (n = 2,118), shape (n = 565). The similarity in results across different question categories reinforces the reliability and stability of our model's performance. For a breakdown of labels per question category, see \autoref{sec:categories}. 




\subsection{GPT-4o Retrieval Prompt}
\begin{framed}
\label{frame:labeling_prompt}
\textbf{system}: Answer the question in one word. Then list the Fact\_ID or Image\_ID of all facts used to derive the answer in square brackets.

\textbf{human}: Question: <query>

\textbf{human}: Text Facts:
[fact\_id\_1: fact\_1, ..., id\_n: fact\_n]

\textbf{human}: Image\_ID: img\_id\_1, \\
Caption: img\_caption\_1

\textbf{human}: [Input\_type=image] \\
image\_url=url\_1

...

\textbf{human}: Image\_ID: img\_id\_m, \\
Caption: img\_caption\_m

\textbf{human}: [Input\_type=image] \\
image\_url=url\_m

\end{framed}

\subsection{Robustness Checks: Question Complexity}
\label{appendix:complexity_analysis_metrics}

\paragraph{Complexity Metrics}
To identify correlations between the complexity of the question, retrieval recall, and QA accuracy, we apply three separate measures to the input questions; Word Count, Flesch-Kincaid Grade Level \citep{flesch2007flesch}, and Gunning-Fog Index \citep{gunning1952technique}.

The Flesch-Kincaid Grade Level is a readability metric that evaluates the difficulty of a text based on the length of its words and sentences \citep{flesch2007flesch}, and is defined as;
\begin{equation}
\begin{split}
  \text{FKGL} = 0.39 \left( \frac{\text{Total Words}}{\text{Total Sentences}} \right) \\ + 11.8 \left( \frac{\text{Total Syllables}}{\text{Total Words}} \right) - 15.59  
\end{split}
\end{equation}

The Gunning Fog Index is a readability test used in linguistics to assess the complexity of English writing \citep{gunning1952technique}, and is defined as;
\begin{equation}
\begin{split}
    \text{GFI} = \frac{0.4 \times \text{Total Words}}{\text{Total Sentences}} \\
    + \frac{40 \times \text{Total Complex Words}}{\text{Total Words}}
\end{split}
\end{equation}

\begin{figure*}
\centering
\includegraphics[width=.3\textwidth]{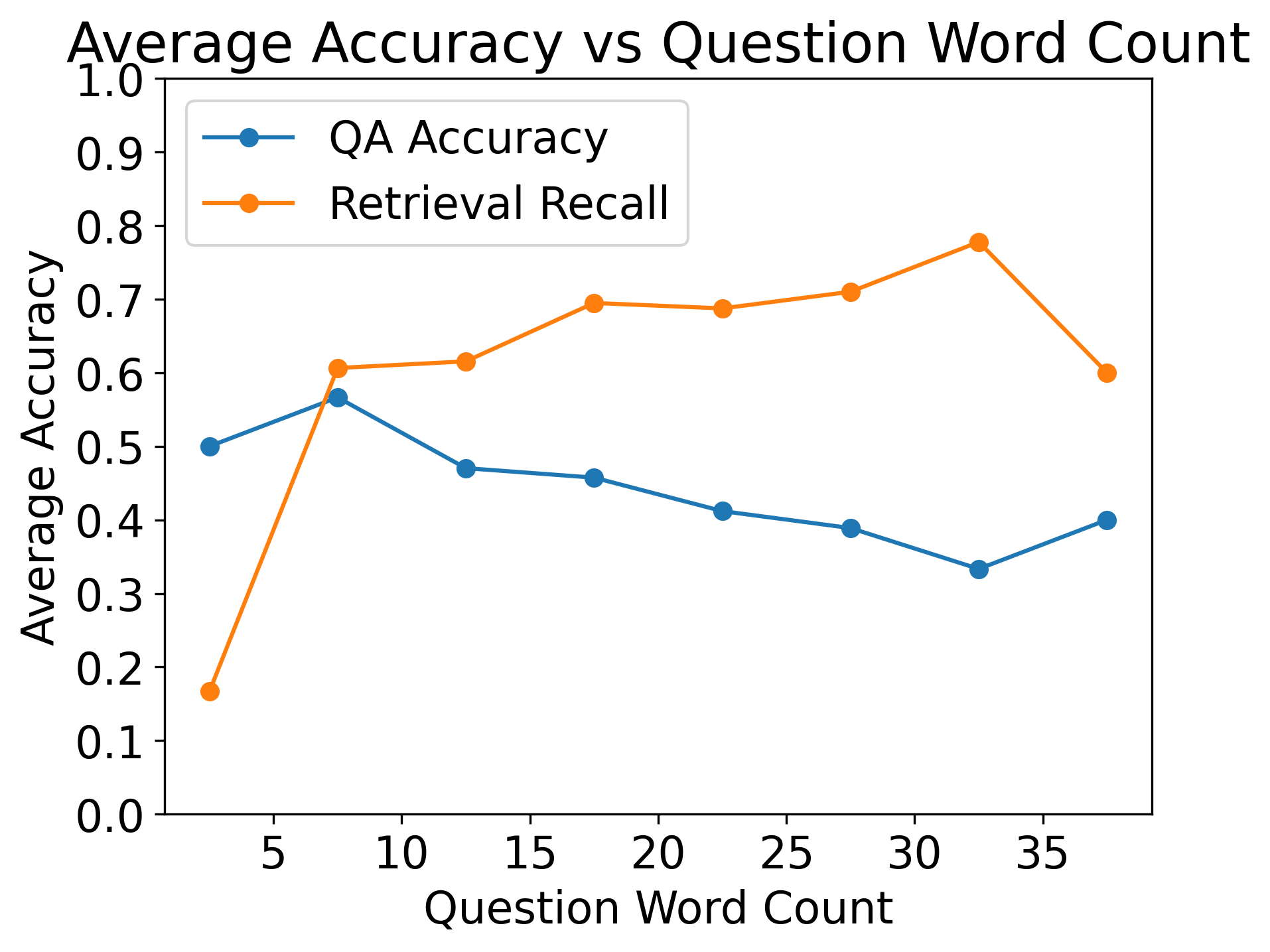}\hfill
\includegraphics[width=.33\textwidth]{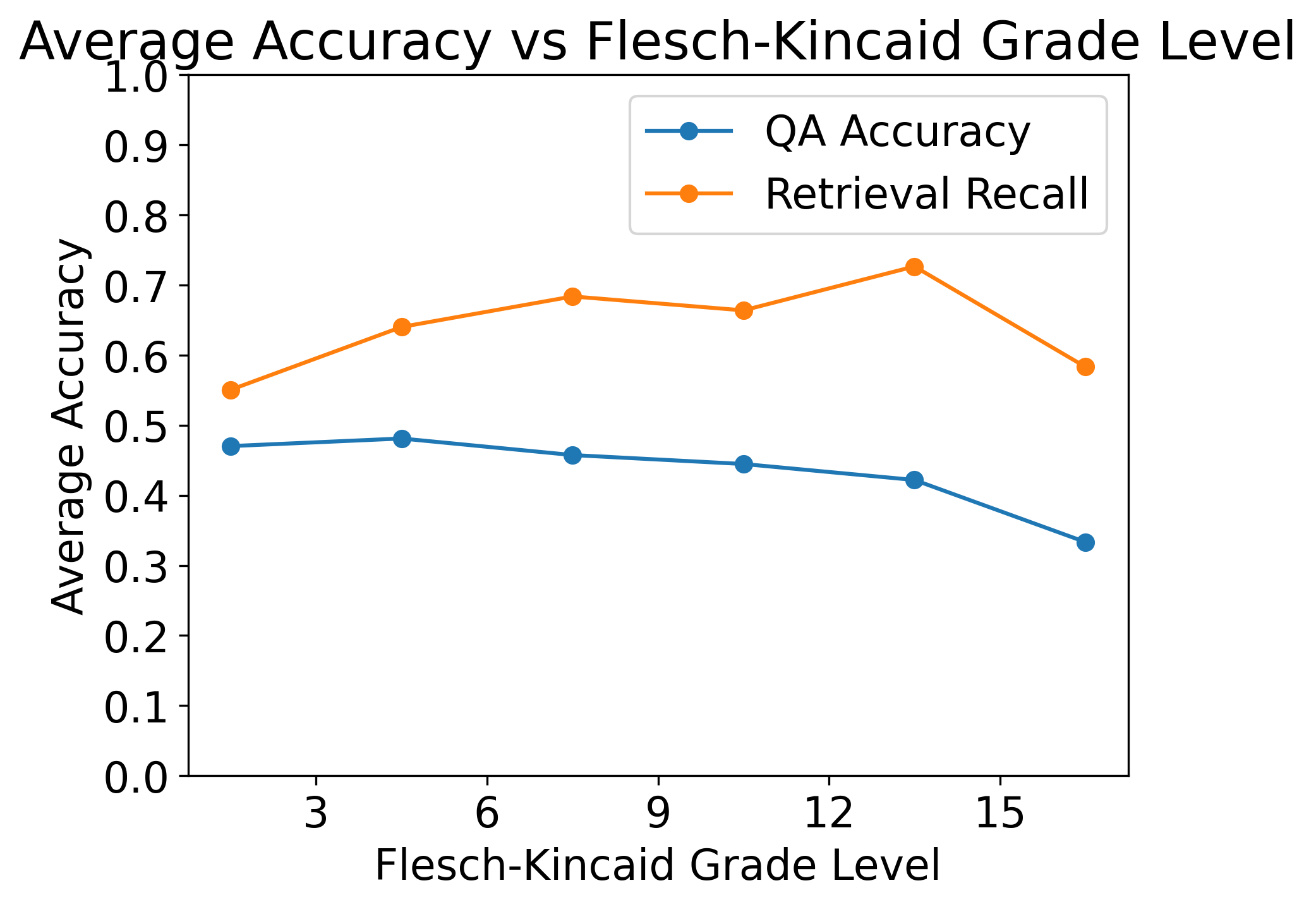}\hfill
\includegraphics[width=.29\textwidth]{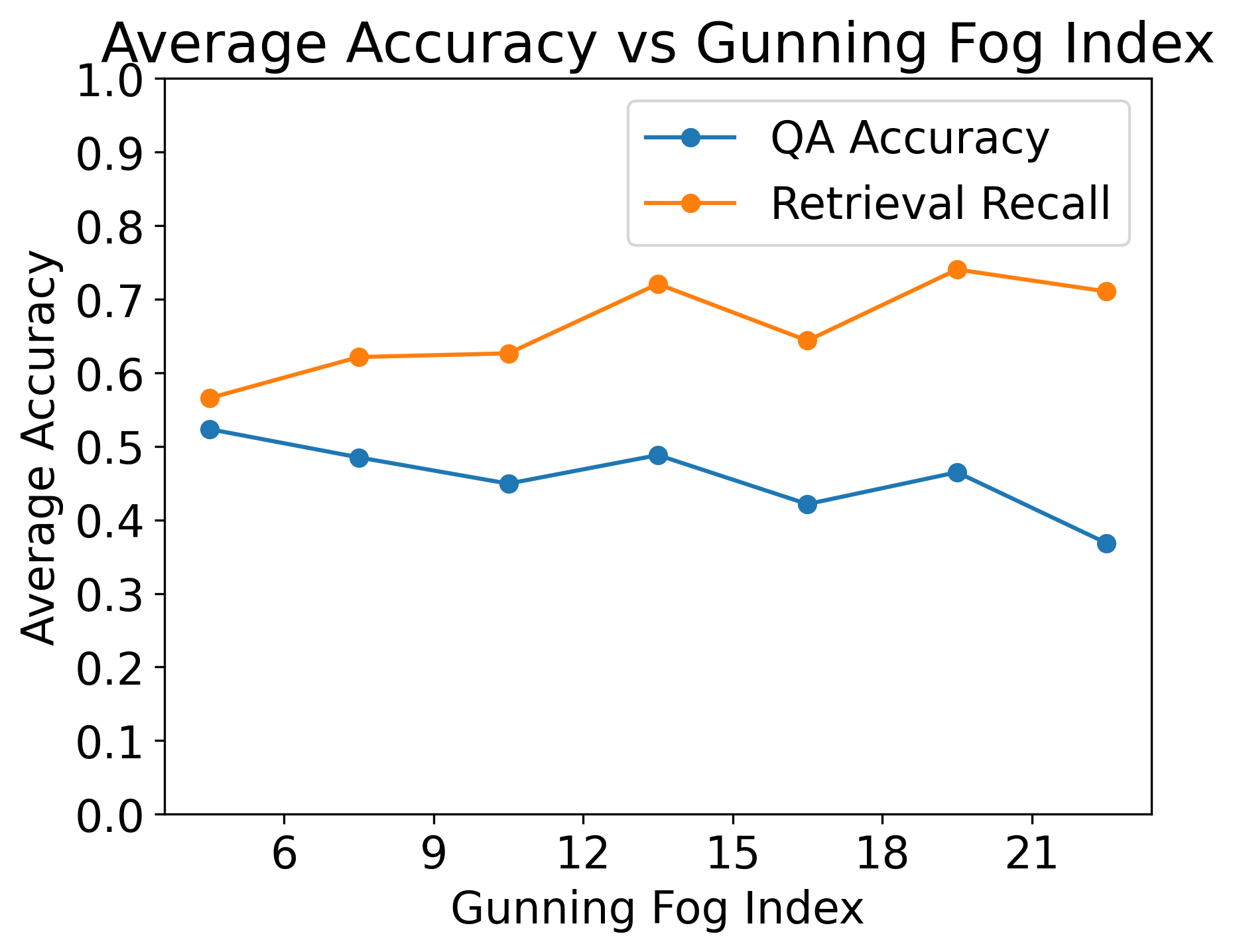}\hfill
    \caption{GPT-4o retrieval and QA performance reveal opposite trends with respect to question complexity; GPT-4o retrieval improves with increased complexity, while GPT-4o QA accuracy degrades.}
    \label{fig:complexity_vs_qa}
\end{figure*}

\begin{figure*}
\centering
\includegraphics[width=.3\textwidth]{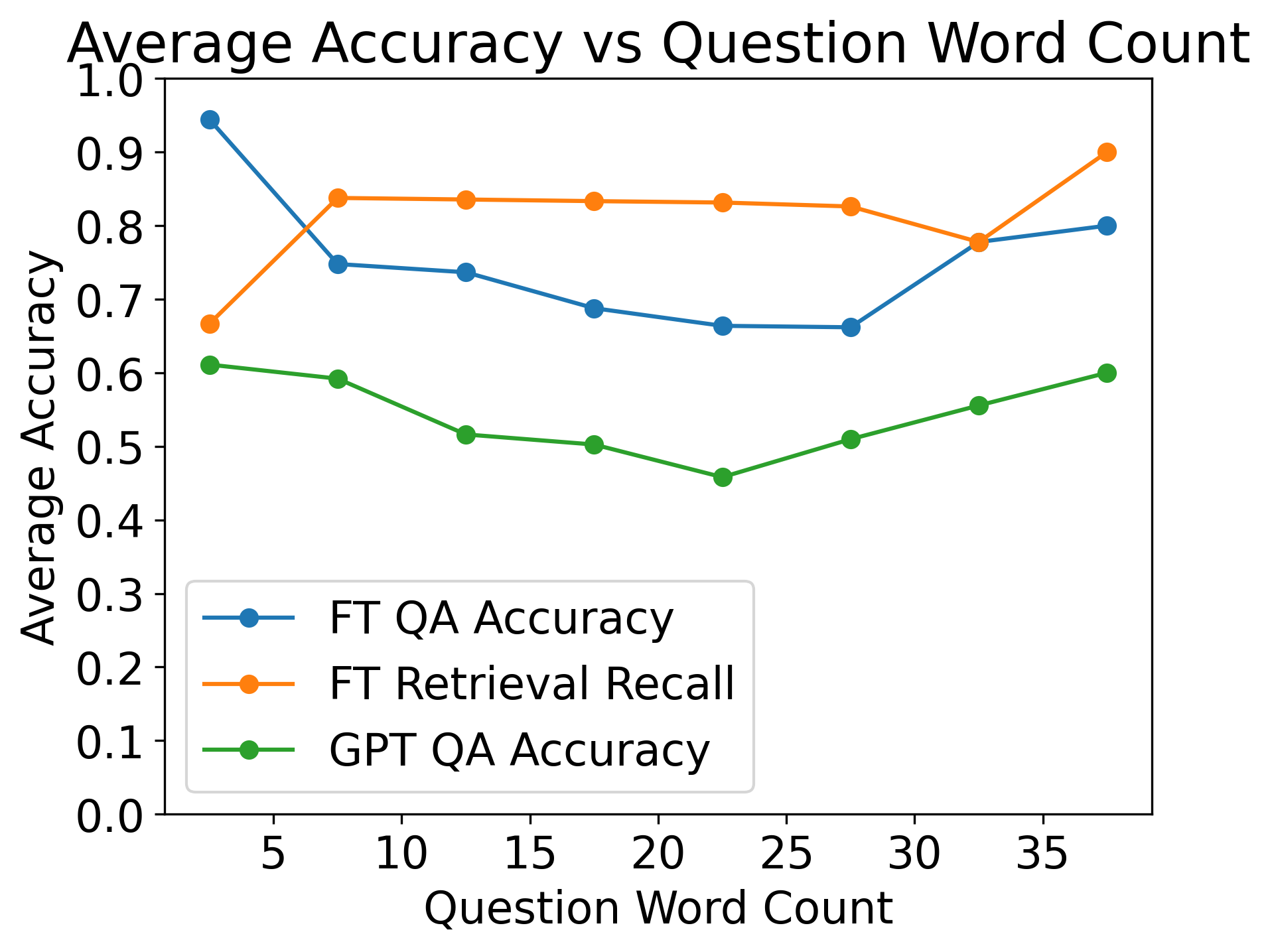}\hfill
\includegraphics[width=.33\textwidth]{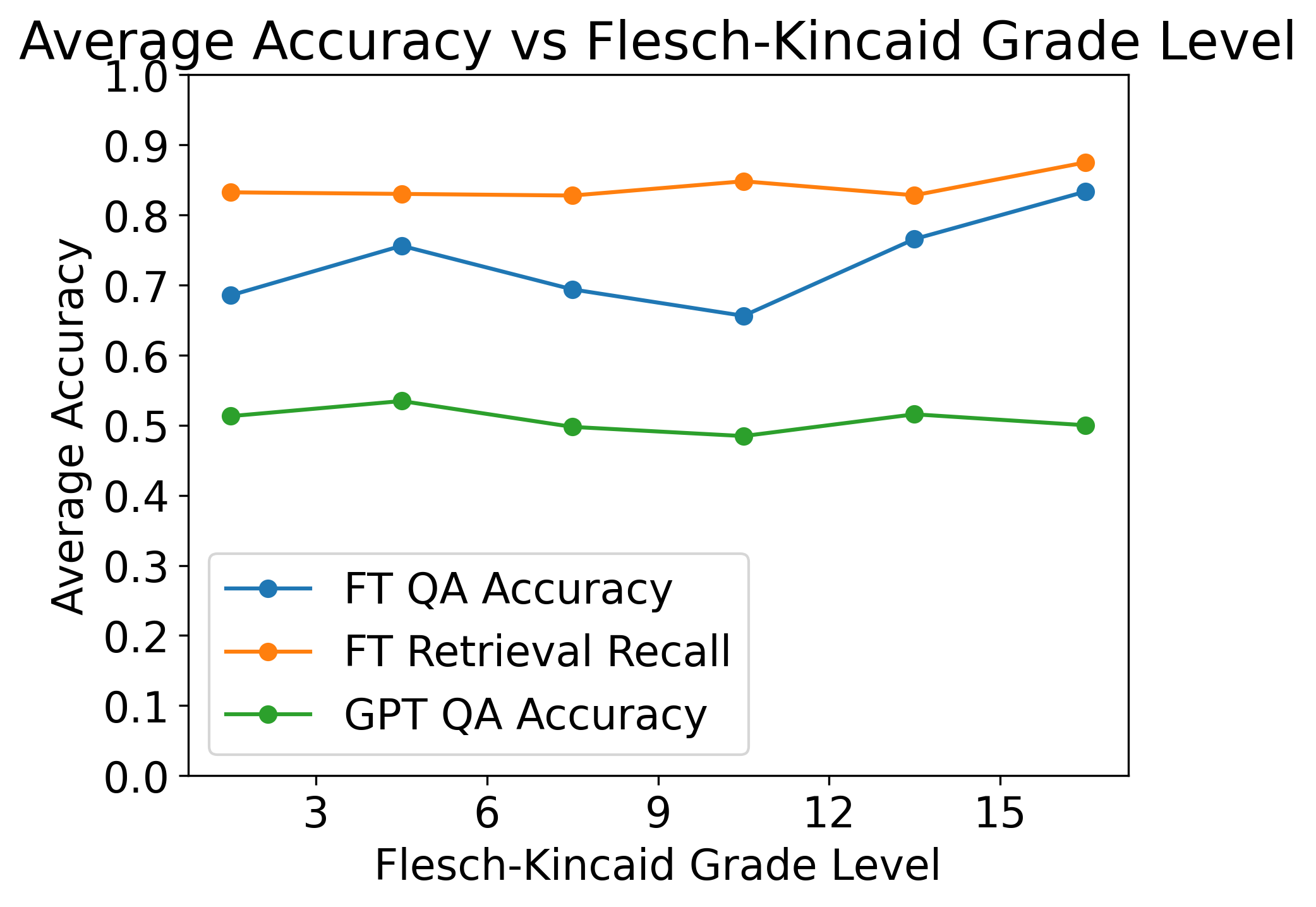}\hfill
\includegraphics[width=.29\textwidth]{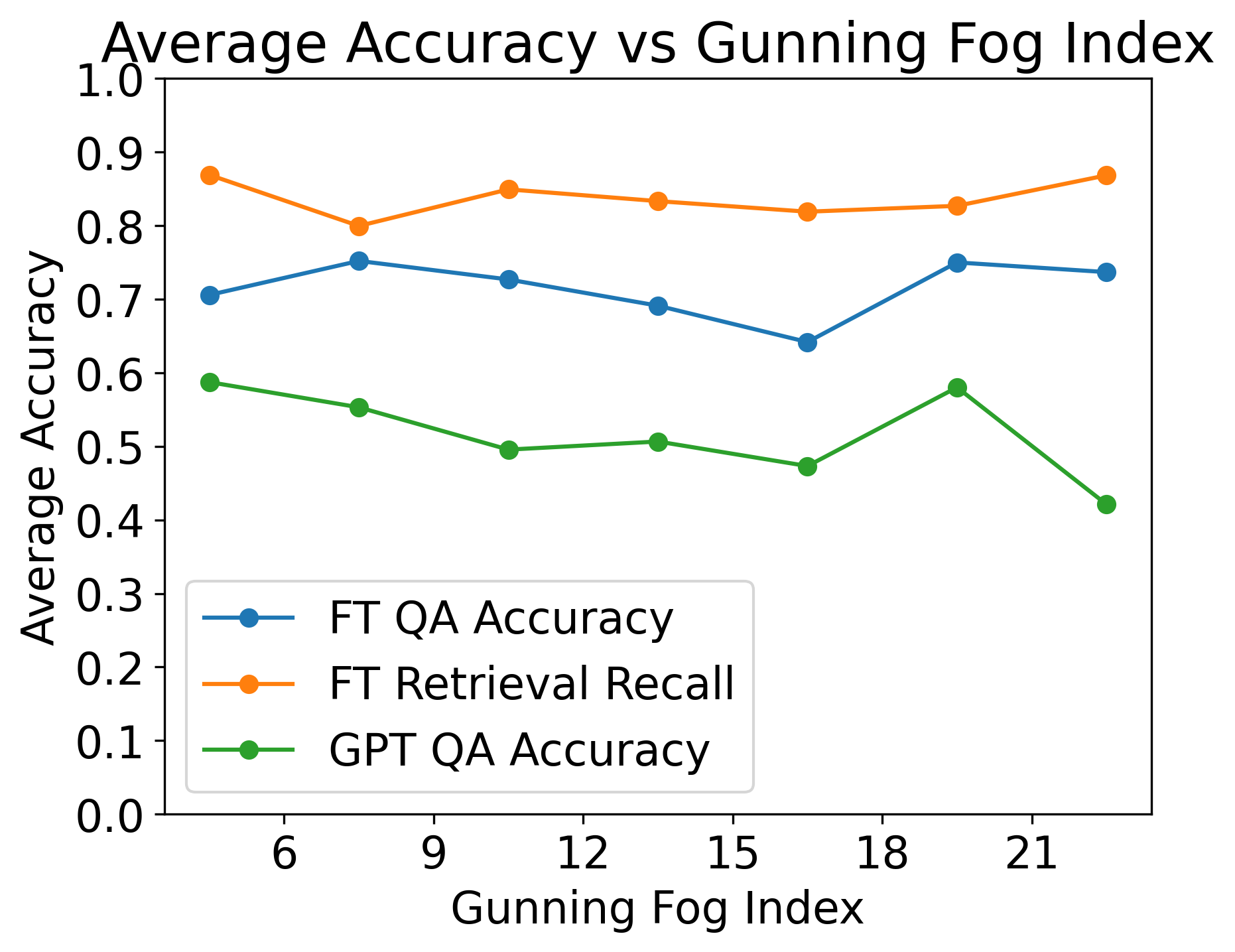}\hfill

    \caption{UniVL-DR retriever performance is independent of question complexity. When coupled with this retriever, the effects of question complexity on GPT-4o and MH-VoLTA QA accuracy is minimized.}
    \label{fig:complexity_vs_qa_ft}
    \vspace{-5mm}
\end{figure*}

\paragraph{Complexity Analysis} 
We observe and report interesting relationships between query complexity and retrieval and QA performance. We find that the accuracy of the in-context GPT-4o retriever is related to question complexity (\autoref{fig:complexity_vs_qa}). The more complex the question in terms of word count, Flesch-Kincaid Grade, or Gunning Fog Index, the lower the QA performance (see \autoref{fig:complexity_vs_qa}). In contrast, increasing query complexity improves GPT-4o's retrieval ability, where the additional complexity provides information on source relevancy. However, this relationship does not hold for the finetuned UniVL-DR retriever, where question complexity has little effect on retrieval recall or QA accuracy (\autoref{fig:complexity_vs_qa_ft}). As such, systems that rely on "in-context" retrieval using GPT-4o are limited by query complexity, but approaches that utilize finetuned retrievers are not.

We note that the opposing relationship between retrieval and QA performance is contrary to the finding that the QA task is heavily dependent on retrieval performance (\autoref{fig:retrieval_vs_qa_lines}). The impact of query complexity on task performance is strong enough to overcome this general principle.

\subsection{Baseline Models}
\label{sec:baselines}

\paragraph{VLP}
\label{sec:VLP}



The VLP transformer model consists of a unified encoder and decoder \citep{zhou_unified_2020}. The VLP architecture is made up of 12 layers of transformer blocks trained according to the BERT bidirectional and the seq2seq objectives where the self-attention module in the transformer block are defined as;

\begin{equation}
    A^l = softmax(\frac{Q^TK}{\sqrt{d}} + M)V^T
\end{equation}

where $V = W^l_VH^l-1, Q = W^l_QH^l-1, K = W^l_KH^l-1$. As in \cite{vaswani2017attention}, a feedforward layer (with residual) maps $A^l$ to $H^l$. The model is trained on image caption pairs, and then finetuned for the VQA task. Finetuning follows by taking the hidden states from the final layer and feeding them to a multi-layer perceptron. The model used has been finetuned twice, once on the VQA dataset (as described by \cite{yu_unified_2023}), and again on the WebQA dataset. 


\paragraph{GIT}
\label{sec:GIT}
To contrast with VLP, a pretrained multihop VQA model, we use a pre-trained Generative Image-to-Text Transformer (GIT) \citep{wang2022git}.
GIT employs a simplified VQA architecture with one encoder for images and one decoder for text. As such, the model is explicitly incapable for multihop VQA between text and images, so it serves as a baseline for pre-trained models that do not utilize image descriptions, and so we concatenate image sources if there are more than one. 


GIT is pre-trained using the language modeling task (as opposed to MLM which is used by VLP) where the model learns to predict captions in an auto-regressive manner. For VQA finetuning, the text input is swapped to the query, so that answers are predicted. 

\paragraph{BLIP-2}
\label{sec:blip}
Similar to VLP, the Bootstrapping Language-Image Pre-training model (BLIP) is a unified vision language pre-trained model \citep{li2022blip}. It relies on a visual transformer which is less computationally demanding and is pre-trained on over 100 Million image-caption pairs using a contrastive loss (ITC) for image-text contrastive alignment and image-text matching (ITM). 
In addition to the ITC and ITM losses, the authors introduce an additional Image-grounded text generation (ITG) loss that trains the Q-former encoder to generate texts, given input images as the condition. 


\paragraph{GPT-3.5 Turbo}
\label{sec:gpt3.5}
Throughout the dataset, a consistent challenge emerges: the model must focus on details, understand them, and accurately respond to questions, even after the provision of positive source images. This challenge has led to the exploration of an image-to-text approach, where the task involves generating descriptive captions for the images. This transforms the problem into a unimodal text retrieval and generation task. Using this method, the SOLAR model has had success on the WebQA task \cite{alibaba_text}. Accordingly, we include a zero-shot oracle baseline, passing queries and image captions to gpt-turbo-3.5 \citep{brown2020language}.

\begin{figure*}
\begin{lstlisting}
yesno_set = {'yes', 'no'}
color_set = {
    'orangebrown', 'spot', 'yellow', 'blue', 'rainbow', 'ivory', 
    'brown', 'gray', 'teal', 'bluewhite', 'orangepurple', 'black', 
    'white', 'gold', 'redorange', 'pink', 'blonde', 'tan', 'turquoise', 
    'grey', 'beige', 'golden', 'orange', 'bronze', 'maroon', 'purple', 
    'bluere', 'red', 'rust', 'violet', 'transparent', 'yes', 'silver', 
    'chrome', 'green', 'aqua'
}
shape_set = {
    'globular', 'octogon', 'ring', 'hoop', 'octagon', 'concave', 'flat', 
    'wavy', 'shamrock', 'cross', 'cylinder', 'cylindrical', 'pentagon', 
    'point', 'pyramidal', 'crescent', 'rectangular', 'hook', 'tube', 
    'cone', 'bell', 'spiral', 'ball', 'convex', 'square', 'arch', 'h', 
    'cuboid', 'step', 'rectangle', 'dot', 'oval', 'circle', 'star', 
    'crosse', 'crest', 'octagonal', 'cube', 'triangle', 'semicircle', 
    'domeshape', 'obelisk', 'corkscrew', 'curve', 'circular', 'xs', 
    'slope', 'pyramid', 'round', 'bow', 'straight', 'triangular', 
    'heart', 'fork', 'teardrop', 'fold', 'curl', 'spherical', 
    'diamond', 'keyhole', 'conical', 'dome', 'sphere', 'bellshaped', 
    'rounded', 'hexagon', 'flower', 'globe', 'torus'
}   
\end{lstlisting}
\caption{WebQA keyword lists per question category.}
\label{sec:categories}
\end{figure*}

\end{document}